\documentclass[10pt,twocolumn,letterpaper]{article}

\usepackage[pagenumbers]{wacv}              

\usepackage{multirow}
\usepackage[table]{xcolor}

\usepackage{graphicx}
\usepackage{tikz}
\usetikzlibrary{arrows.meta}

\newcommand{\gnode}[2]{%
  \begin{tikzpicture}[remember picture, inner sep=0pt]
    \node (#1) {\includegraphics[width=\linewidth]{#2}};
  \end{tikzpicture}%
}

\usepackage{diagbox}

\definecolor{wacvblue}{rgb}{0.21,0.49,0.74}
\usepackage[pagebackref,breaklinks,colorlinks,allcolors=wacvblue]{hyperref}

\def\wacvPaperID{721} 
\def\confName{WACV}
\def\confYear{2027}

\title{Where to Cut? Adaptive Thresholds for Density-Ratio Few-Shot Segmentation}

\author{Rafael Sterzinger\thanks{Recipient of a DOC Fellowship of the Austrian Academy of Sciences.} \qquad Robert Sablatnig\\
Computer Vision Lab, TU Wien\\
Vienna, Austria\\
{\tt\small \{firstname.lastname\}@tuwien.ac.at}
}
\begin{document}
\maketitle

\begin{abstract}
A deployed system meets new parts, objects, and domains faster than they can be labeled and trained for.
Naturally, few-shot segmentation lends itself as a solution: the target is specified at inference time from a few annotated images instead.
Since the emergence of vision foundation models, frozen features carry this setting untrained, leaving one question: how to turn them into a mask.
INSID3 partitions the query, fixing the mask's granularity before the support is seen; FROST drops the partition and scores every token by a density ratio, reaching the resolution of the feature grid, only to fix the decision at the Bayes zero of an equal class prior.
We measure what that constant costs: on an unchanged score field, the best per-episode cut is worth 14 mIoU more at one shot, and the margin grows with the support --- a threshold calibrated on the support does not transfer to the query.
We therefore read the cut from the query itself: a valley separator on the histogram of the score's positive part, corrected by shrinkage, with no threshold constant to tune.
Across nine benchmarks spanning semantic, part, and personalized segmentation, we lead on all, reaching 58.6 mIoU (+3.9) at one shot; over FROST the gains are largest where its equal-prior assumption is furthest off (+10.6 COCO-20$^i$, +10.0 LVIS-92$^i$).
At five shots, we reach 66.0 mIoU (+2.3) and lead from one to ten shots; adaptive thresholding alone accounts for +7.2 / +8.5 of the gain.
\end{abstract}

\section{Introduction}
Acquiring dense mask annotations for visual scene understanding remains challenging, as no fixed label set can cover what a deployed system might observe: a new part~\cite{liu2024matcher}, a new object~\cite{liu2024matcher}, or a new domain --- aerial~\cite{yao2022scale}, underwater~\cite{islam2020semantic}, medical~\cite{candemir2014lung,codella2018skin} or historical documents~\cite{sterzinger2025map,sterzinger2025text} --- arrives faster than it can be labeled, and retraining for each is neither affordable nor timely.
A segmentation task, therefore, has to be specifiable at inference time, from a few examples rather than a training set.
In scientific terms, we refer to this setting as few-shot segmentation (FSS), or in-context segmentation: given $k$ annotated support images and a query image, FSS asks for the mask of a concept the model has never been trained on.

\begin{figure}[t]
    \centering
    \begin{subfigure}{0.32\columnwidth}
        \centering
        \includegraphics[width=\linewidth]{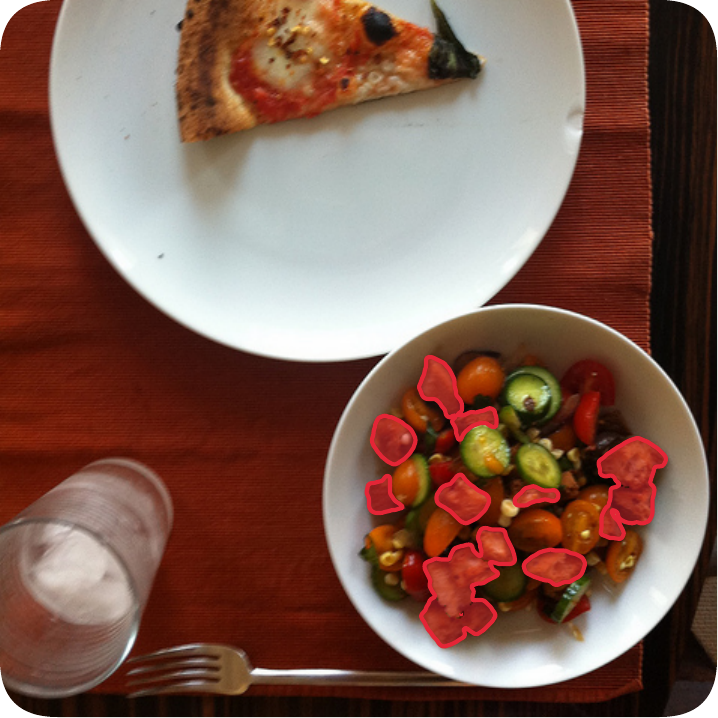}
        \caption{Query}
        \label{fig:sub2}
    \end{subfigure}
    \hfill
    \begin{subfigure}{0.32\columnwidth}
        \centering
        \includegraphics[width=\linewidth]{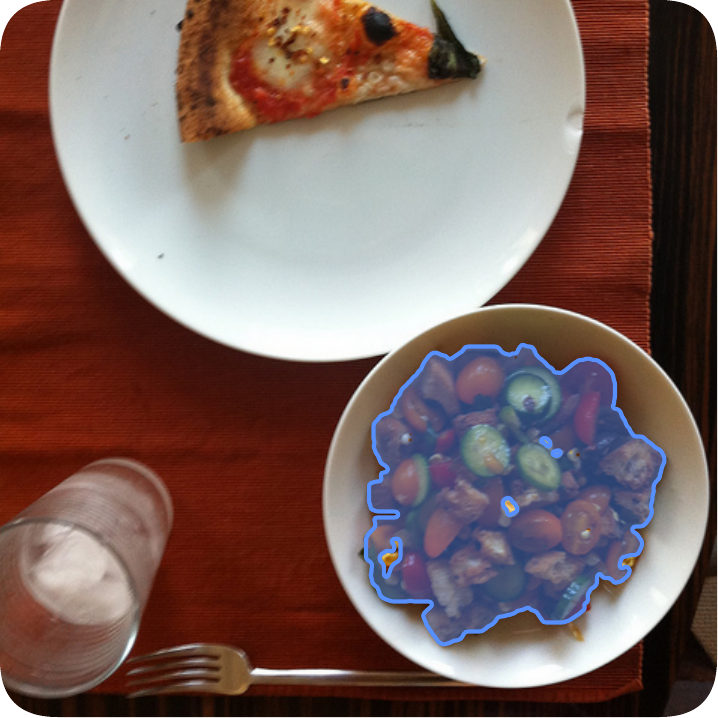}
        \caption{INSID3~\cite{cuttano2026insid3}}
        \label{fig:insid3}
    \end{subfigure}
    \hfill
    \begin{subfigure}{0.32\columnwidth}
        \centering
        \includegraphics[width=\linewidth]{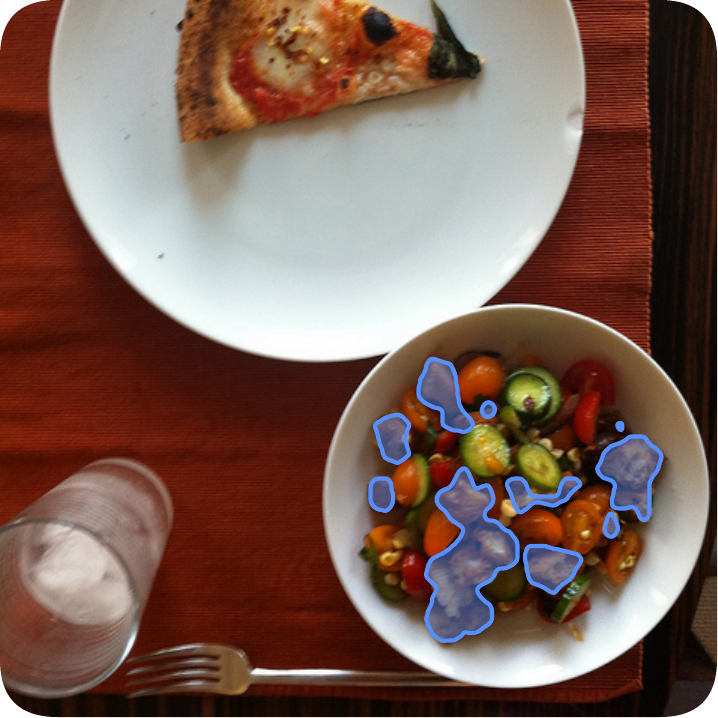}
        \caption{Ours}
        \label{fig:ours}
    \end{subfigure}

    \caption{\textbf{Teaser.} A one-shot LVIS-92$^i$ episode: query w/ ground truth (a), INSID3~\cite{cuttano2026insid3} (b, IoU $28$), ours (c, IoU $70$).
    INSID3's partition is fixed before support is seen and has no finer clusters than the bowl; KDE-based density-ratio scoring recovers the pieces.}
    \label{fig:teaser}
\end{figure}

\begin{figure*}[t]
    \centering
    \tikzset{flow/.style={-{Stealth[length=2.5mm,width=2mm]},
                          line width=0.9pt, draw=black!55}}

    \begin{subfigure}{0.22\textwidth}
        \centering
        \gnode{pa}{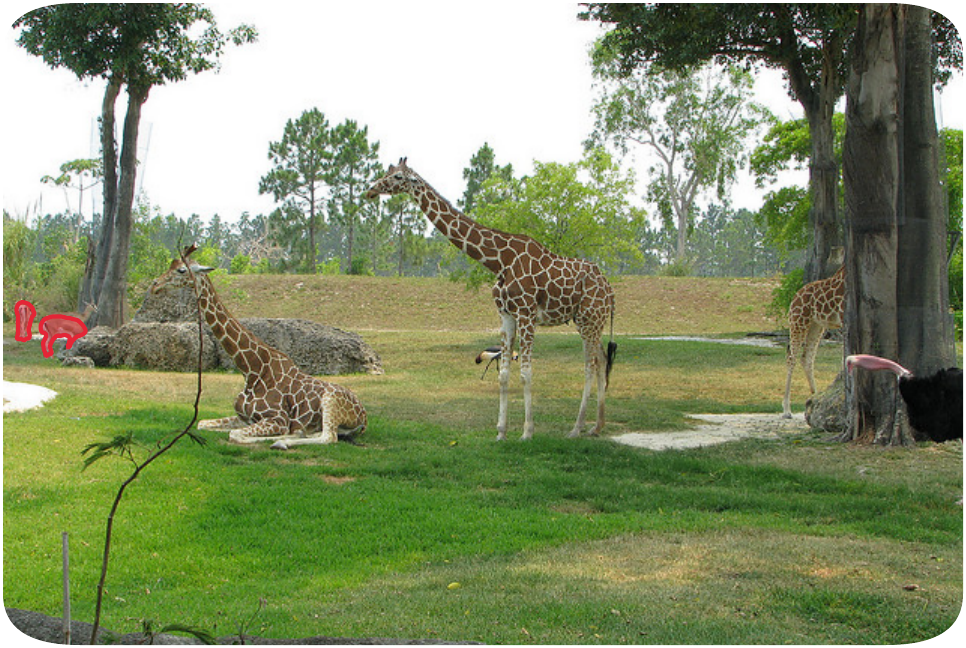}
        \caption{Support + Ground Truth}
    \end{subfigure}\hfill
    \begin{subfigure}{0.22\textwidth}
        \centering
        \gnode{pb}{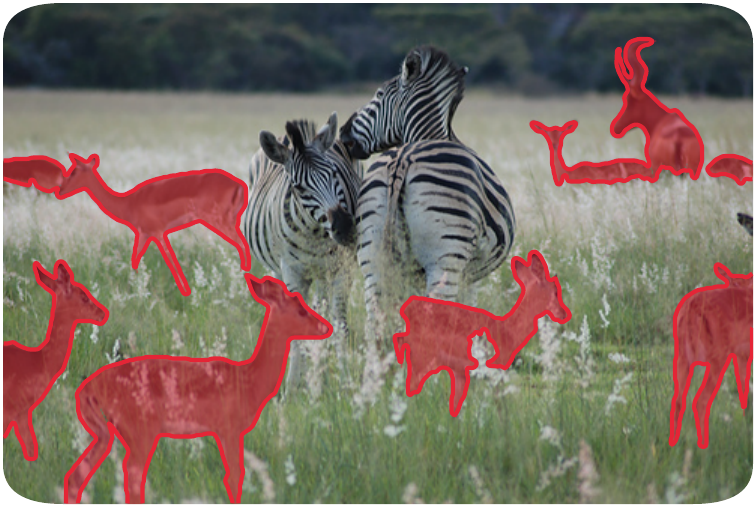}
        \caption{Query + Ground Truth}
    \end{subfigure}\hfill
    \begin{subfigure}{0.22\textwidth}
        \centering
        \gnode{pc}{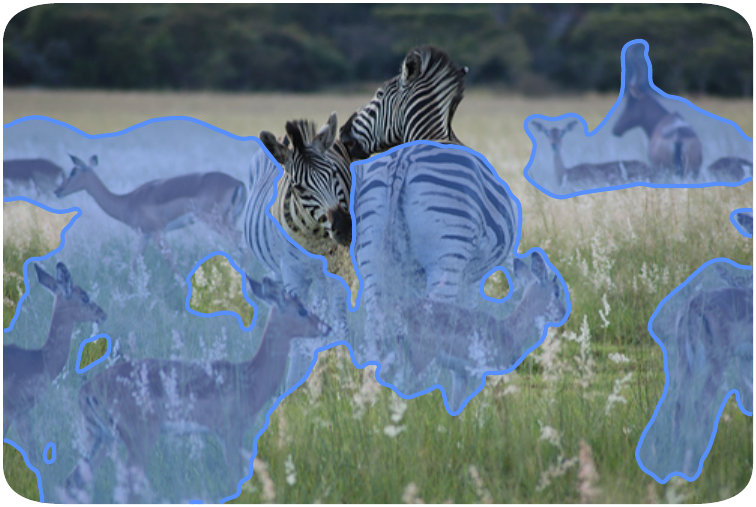}
        \caption{Mask at $\tau_0$}
    \end{subfigure}\hfill
    \begin{subfigure}{0.22\textwidth}
        \centering
        \gnode{pd}{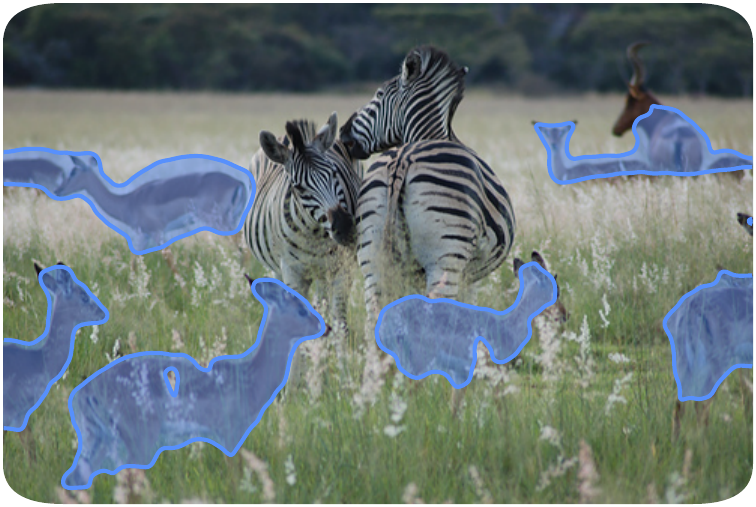}
        \caption{Mask at $\hat{\tau}$}
    \end{subfigure}

    \vspace{0.45cm}
    \begin{subfigure}{0.22\textwidth}
        \centering
        \gnode{pe}{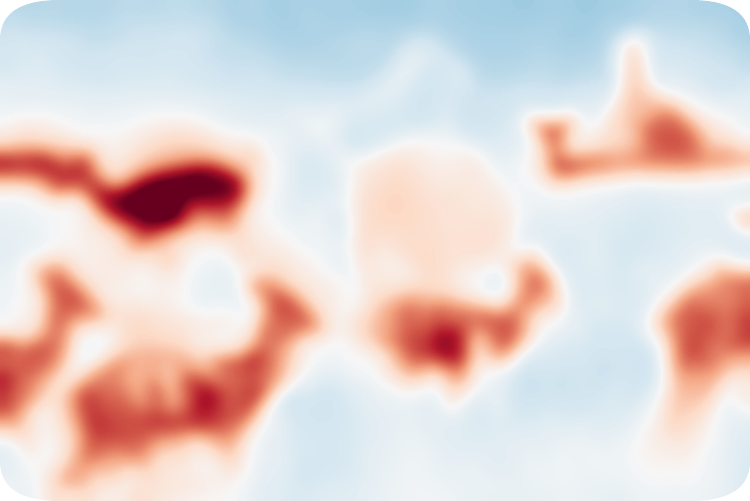}
        \caption{Step 0: 2D Score Field}
    \end{subfigure}\hfill
    \begin{subfigure}{0.22\textwidth}
        \centering
        \gnode{pf}{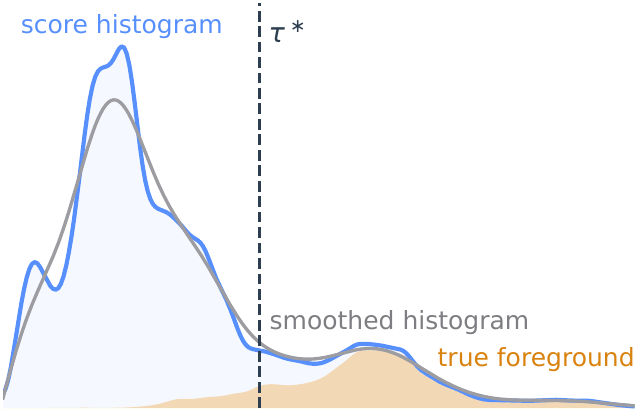}
        \caption{Step 1: 1D Score Field}
    \end{subfigure}\hfill
    \begin{subfigure}{0.22\textwidth}
        \centering
        \gnode{pg}{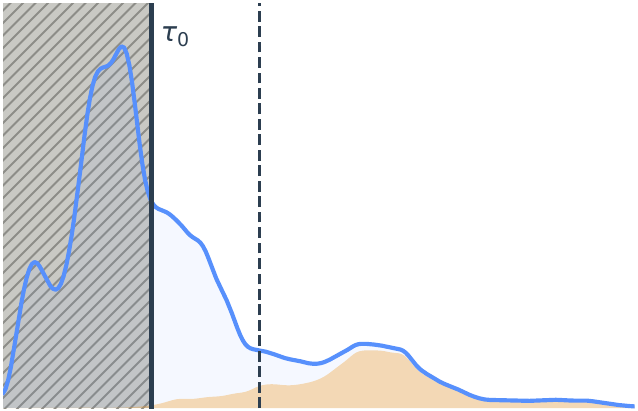}
        \caption{Step 2: Applying Window}
    \end{subfigure}\hfill
    \begin{subfigure}{0.22\textwidth}
        \centering
        \gnode{ph}{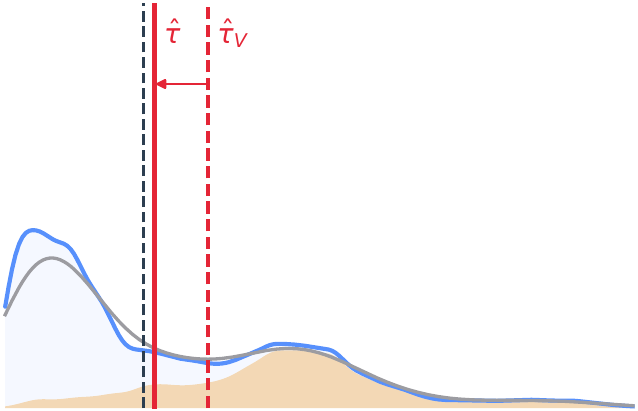}
        \caption{Step 3: Predicting $\hat{\tau}$}
    \end{subfigure}

    \begin{tikzpicture}[remember picture, overlay]
      \draw[flow] ([xshift=1mm,yshift=-0.4mm]pe.east) -- (pf.west);
      \draw[flow] ([xshift=1mm]pf.east) -- (pg.west);
      \draw[flow] ([xshift=1mm]pg.east) -- (ph.west);
      \draw[flow] ([yshift=1mm]pg.north) -- ([yshift=-4mm]pc.south);
      \draw[flow] ([yshift=1mm]ph.north) -- ([yshift=-4mm]pd.south);
    \end{tikzpicture}

    \caption{\textbf{Overview.} Given a support image with its mask and a query, we score every query token by $s(q)=\ell_\sigma(q)+\lambda\,\mathrm{fwd}(q)$, the KDE log-ratio plus a support$\to$query correspondence term. At $\tau_0=0$, the cut lands inside the background mode and here segments the whole herd. We instead read the cut from the query score distribution: the histogram is restricted to the window $\mathcal{R}=\{s>0\}$, its valley $\hat\tau_V$ separates the two modes, and the shrinkage $\hat\tau=\hat\tau_V/(1+\hat\pi)$ moves the cut toward $\tau^\ast$ (dashed, gray); blue: prediction, red: ground truth.}
    \label{fig:overview}
\end{figure*}

Since the rise of vision foundation models~(VFMs)~\cite{kirillov2023sam,oquab2024dinov,simoni2025dinov3}, the support examples no longer have to teach the features, i.e., dense self-supervised tokens already relate a query to its support and delineate objects within it, and the open question has become how to turn them into a mask.
For instance, decoder-based methods~\cite{liu2024sine,meng2024segic} train a segmentation head on frozen DINOv2 features, which works well in-domain but requires a training set and does not generalize; matching-based methods~\cite{liu2024matcher,zhang2024personalize} pair the VFM's correspondences with a promptable segmenter such as SAM~\cite{kirillov2023sam}, avoiding training at the cost of a second model with a segmentation prior of its own.
INSID3~\cite{cuttano2026insid3} showed that frozen DINOv3 features alone suffice to replace both: the query is clustered agglomeratively, the cluster most similar to a support prototype is selected, and affine neighbors are merged into it.
The partition is a strong prior on object-centric imagery, but its granularity is fixed before the support is consulted, so a concept finer than the clusters is out of reach (cf.\ Fig.~\ref{fig:teaser}; \ref{fig:insid3} vs. \ref{fig:ours}), and the merge requires a cut on cluster affinity --- two a-priori tuned constants.
Subsequently, FROST~\cite{park2026frost} shows that the partition can be omitted altogether: foreground and background support tokens define two kernel densities, each query token is scored by its log ratio, and the mask is cut where the ratio changes sign; the Bayes threshold under an equal class prior.
Consequently, the decision is per token, limited only by the feature grid, and free of tuned cuts.
On the downside, it presumes that a score fitted on the support keeps its scale on the query.

In this work, we show that on FROST's score field, the fixed cut leaves $14$ mIoU to the best per-episode oracle cut at one shot, and the loss grows with the number of shots, as more examples sharpen the ranking that a fixed cut cannot exploit.
We therefore ask whether a better cut can be read off the query itself: a per-episode threshold, computed solely from the score distribution, that closes this gap without requiring a constant to be tuned (\Cref{fig:overview}).
Our answer rests on one observation: a classical histogram separator finds the right cut when it sees the right histogram: over the whole image, the score is one background bulk with a foreground tail, and any separator lands inside the bulk; over the tokens, the score itself calls foreground, two modes stand apart, and their valley lies close to the oracle cut.
In summary, we make the following contributions:
\begin{itemize}
  \item \textbf{A diagnosis of the fixed cut.} Correcting FROST's prior
  assumption with the true foreground fraction makes the cut worse, while
  keeping the same fraction of the query by rank recovers the oracle within
  $0.4$ mIoU. The ranking is sound and the zero is not: the failure is one of
  calibration, not of prior, and a rank-based cut is immune to it.
  \item \textbf{An episode-adaptive threshold rule.} On FROST's density-ratio
  histogram augmented with a correspondence term, restricted to likely-foreground
  patches, we place the cut at the valley between its two modes and pull it back
  by a scale-invariant correction; a backward correspondence then removes what
  the support does not reciprocate. The three surviving constants obey a single
  scaling law, so only a temperature and one dimensionless ratio are selected.
  \item \textbf{State-of-the-art training-free results.} On nine benchmarks
  spanning semantic, part and personalized segmentation, we reach $58.6$ mIoU at one shot,
  $+4.8$ over FROST and ahead on every benchmark, and $66.0$ at five shots
  ($+2.3$), holding the lead from one to ten shots. The adaptive cut is the
  largest single step of the ablation ($+7.2$ / $+8.5$ over the fixed cut at
  one / five shots).
\end{itemize}
To foster reproducibility, we make our code publicly available at: \href{https://github.com/RafaelSterzinger/adaptive-cut}{github.com/RafaelSterzinger/adaptive-cut}.

\section{Related Work}

\paragraph{Few-shot and in-context segmentation.}
Classically, few-shot segmentation is learned from episodes over base classes and evaluated on disjoint new ones, either by matching query pixels against prototypes of the support foreground~\cite{wang2019panet,tian2022prior} or by aggregating dense support--query correlations~\cite{min2021hsnet}; more recently by fitting a mask decoder over a frozen foundation backbone~\cite{meng2024segic,liu2024sine}.
A second group remains training-free and instead prompts a segmenter such as SAM~\cite{kirillov2023sam}, reading prompts from support--query correspondences~\cite{liu2024matcher,zhang2024personalize}, selecting them by a graph over candidate points and masks~\cite{zhang2024bridge}, or replacing them by matching against a memory of reference features~\cite{espinosa2025notime}; such pipelines inherit the granularity and cost of the segmenter they call.
A third group uses a single frozen backbone; self-supervised ViTs localise objects without labels~\cite{caron2021emergent}, a property sharpened by patch-level objectives~\cite{zhou2021ibot} and by scale~\cite{oquab2024dinov,simoni2025dinov3}, and three recent methods read a mask directly from these features: INSID3~\cite{cuttano2026insid3} clusters the query and selects the region matching the reference, FSSDINO~\cite{zakir2026fssdino} matches support prototypes with a Gram-matrix refinement, and FROST~\cite{park2026frost} assigns every query token the ratio of two kernel densities fitted to the foreground and background support anchors.
The step that turns these scores into a mask is a constant fixed a-priori: a cut on cluster affinity in INSID3, the Bayes zero under an equal class prior in FROST.

\paragraph{Adaptive thresholding.}
Cutting a histogram is a classical problem. Otsu's method~\cite{otsu1979otsu} maximizes the between-class variance, minimum-error thresholding~\cite{kittler1986minimum} fits a two-Gaussian mixture, and the \textit{minimum} method~\cite{prewitt1966valley} smooths the histogram until two modes remain and cuts at the valley; Glasbey~\cite{glasbey1993analysis} compares these estimators.
For the metric at hand, the expected-IoU objective of Nowozin~\cite{nowozin2014optimal} implies that the optimal decision on calibrated posteriors is not the equal-posterior point but a lower one, set by the achievable IoU.
However, such rules are rarely used in few-shot segmentation; Herzog's ABCDFSS~\cite{herzog2024adapt} binarises its fused correlation map at the larger of the map's mean and Otsu's cut, in place of a tuned constant.
A cut can also be derived from the foreground fraction of the query, which is the subject of class-prior estimation~\cite{saerens2002adjusting} and positive--unlabelled learning~\cite{elkan2008learning}.
We take the first route, estimating the cut per episode from the shape of the query score distribution.
\begin{table}[t]
\centering
\scalebox{0.95}{
\begin{tabular}{llc}
        \toprule
        Cut rule & $\tau$ & Avg. \\
        \midrule
        \rowcolor{gray!15}
        Equal-prior Bayes& $0$ & 52.4 \\ 
        \midrule
        True-prior Bayes & $\log\!\big((1-\pi^\ast)/\pi^\ast\big)$
            & 51.5 \\
        True-prior quantile & $Q_{1-\pi^\ast}(\ell)$ & \underline{66.2} \\
        \rowcolor{red!10}
        Oracle cut & $\tau^\ast$ & \textbf{66.6} \\
        \bottomrule
\end{tabular}
}
\caption{\textbf{Oracle analysis.} FROST's threshold against oracle headroom; mIoU averaged over the full benchmark at $k=1$. The equal-prior row reproduces~\cite{park2026frost}; $\pi^\ast$ is the true query foreground fraction and $\tau^\ast$ the per-episode IoU-optimal cut.}
\label{tab:benchmark_comparison}
\end{table}

\section{Preliminaries and Motivating Analysis} \label{sec:preliminary}
In this section, we first review the preliminaries of FROST's~\cite{park2026frost} scoring pipeline, from kernel density estimation over patch embeddings to the resulting log-ratio score.
We then analyze the Bayes cut-off $\tau_0$ implied by an equal class prior, show where and why this fixed threshold fails, and motivate replacing it with an adaptive one.

\subsection{Kernel Density Estimation}\label{sec:preliminaries:kde}
FROST~\cite{park2026frost} scores query patches against the support set with a
kernel density estimate~(KDE) over DINOv3~\cite{simoni2025dinov3} patch
embeddings, $\ell_2$-normalized, denoised following
INSID3~\cite{cuttano2026insid3}, and whitened with a heavily shrunk
within-class scatter estimate from the support (see~\cite{park2026frost}).

Let $\mathcal{A}_c$ be the support anchors of class
$c\in\{\mathrm{FG},\mathrm{BG}\}$, with $N_c=|\mathcal{A}_c|$,
$N_{\mathrm{S}}=N_{\mathrm{FG}}+N_{\mathrm{BG}}$, and $N_{\mathrm{Q}}$ the
number of query patches. For a query embedding $q$, each class-conditional
density is a von Mises--Fisher (vMF) KDE with a bandwidth $\sigma$ shared across
classes,
\begin{equation}
  \hat p_c(q;\sigma)=\frac{1}{N_c}\sum_{a\in\mathcal{A}_c}\frac{1}{Z_\sigma}
  \exp\Big(\frac{\cos(q,a)}{\sigma}\Big),
  \label{eq:kde}
\end{equation}
with $Z_\sigma$ the normalizing constant, and each patch is scored by the log
ratio
\begin{equation}
  \ell_\sigma(q)=\log \hat p_{\mathrm{FG}}(q;\sigma)
                -\log \hat p_{\mathrm{BG}}(q;\sigma),
  \label{eq:llr}
\end{equation}
in which $Z_\sigma$ cancels. Under an equal class prior, Bayes' rule places the
decision threshold at $\tau_0=0$.

\paragraph{Bandwidth selection.}
FROST fixes $\sigma$ from the support alone, before any query is scored: over a
small grid it maximizes the leave-one-out (LOO) class margin
\begin{equation}
  M(\sigma)=\frac{1}{N_{\mathrm{S}}}
  \sum_{a\in\mathcal{A}_{\mathrm{FG}}\cup\mathcal{A}_{\mathrm{BG}}}
  s_a\, \ell_\sigma^{\setminus a}(a),
  \label{eq:loo_margin}
\end{equation}
where $s_a=\pm1$ for foreground and background anchors and
$\ell_\sigma^{\setminus a}(a)$ is \Cref{eq:llr} evaluated at $a$ with $a$ removed.

\begin{figure*}[t]
    \centering
    \includegraphics[width=0.78\textwidth]{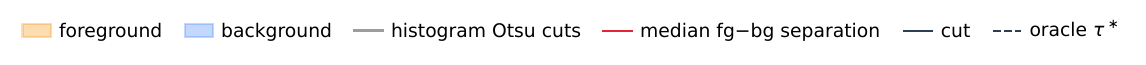}
    \vspace{-0.5cm}
\end{figure*}

\begin{figure}[t]
    \centering
    \begin{subfigure}[t]{0.42\linewidth}
        \centering
        \includegraphics[width=\linewidth]{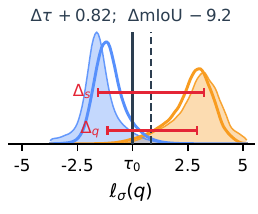}
        \caption{X-Ray, $c=0.84$}
        \label{fig:contraction-xray}
    \end{subfigure}
    \hspace{1cm}
    \begin{subfigure}[t]{0.42\linewidth}
        \centering
        \includegraphics[width=\linewidth]{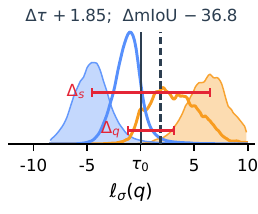}
        \caption{LVIS-92$^i$, $c=0.39$}
        \label{fig:contraction-lvis}
    \end{subfigure}
\caption{\textbf{Contraction.} $\ell_\sigma(q)$ \cref{eq:llr} fitted on the support (filled: LOO anchors) but applied to the query (outlined), on a common scale; $c=\Delta_q/\Delta_s$. $\Delta\tau$: distance from $\tau_0$ to the median oracle $\tau^\ast$; $\Delta$mIoU: median per-episode IoU lost at $\tau_0$. Where the separation survives (a), $\tau_0$ sits near $\tau^\ast$; where it collapses (b), it falls far left.}
    \label{fig:contraction}
\end{figure}

\subsection{Oracle Analysis}
The equal-prior assumption behind $\tau_0$ is not innocuous: object-centric
datasets have markedly smaller foreground fractions than the overhead imagery
for which it is plausible, suggesting that prior mismatch explains
FROST's~\cite{park2026frost} weaker results on LVIS-92$^i$ and COCO-20$^i$. We test this against oracle variants with access to the true query foreground fraction $\pi^\ast$ (\Cref{tab:benchmark_comparison}).

It does not: the Bayes-optimal correction $\log\!\big((1-\pi^\ast)/\pi^\ast\big)$
\emph{worsens} performance, $52.4 \to 51.5$~mIoU, while the same prior applied
as a quantile cut --- keeping the top-$\pi^\ast$ fraction of patches by score
--- yields $66.2$, within $0.4$ of the per-episode optimal cut $\tau^\ast$; supplying the true prior does not help; using it as a rank does.

The failure is therefore one of calibration: the KDE is fitted to the support, but its log-ratio scale does not carry over to the query.
\Cref{fig:contraction} makes this visible. On LVIS-92$^i$ (\ref{fig:contraction-lvis}) the median foreground--background separation contracts from $\Delta_s = 11.2$ to $\Delta_q = 4.4$, and does so \emph{asymmetrically}, so $\tau_0$ lands off-centre in the query score field and costs $36.8$~mIoU against $\tau^\ast$.
On X-Ray~(\ref{fig:contraction-xray}), where the equal-prior regime roughly holds, the separation survives ($4.8$ vs.\ $4.1$) and the loss falls to $9.2$.
A rank-based cut is immune to exactly this: monotone distortions of the scale leave quantiles intact.

\subsection{Threshold Prediction}
The oracle quantile is unavailable at test time, as it requires $\pi^\ast$; the cut must therefore be estimated from the query's score distribution alone.
For this, we consider Otsu's method~\cite{otsu1979otsu} as a baseline, which~\citet{herzog2024adapt} also adopts for its correlation maps.
We observe that Otsu succeeds or fails depending on the \emph{support} of the histogram it sees: applied to the full score field, it is dominated by the background mode; applied to a gated candidate region, it recovers a near-optimal cut.

\begin{figure}[t]
    \centering
    \begin{subfigure}[t]{0.42\linewidth}
        \centering
        \includegraphics[width=\linewidth]{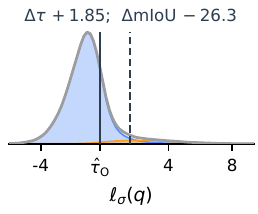}
        \caption{Full}
        \label{fig:full-lvis}
    \end{subfigure}
    \hspace{1cm}
    \begin{subfigure}[t]{0.42\linewidth}
        \centering
        \includegraphics[width=\linewidth]{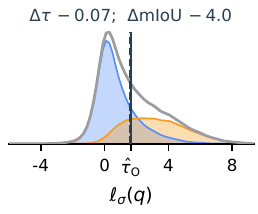}
        \caption{Gated}
        \label{fig:gated-lvis}
    \end{subfigure}
    \caption{\textbf{Gating.} Annotated as in \Cref{fig:contraction}, with $\hat\tau_{O}$ as the cut in use: (a) ungated, foreground carries a few percent of the pixel mass and $\hat\tau_O$ lands inside the background mode; (b) the candidate gate removes $96\%$ of background against $9\%$ of foreground, lifting the foreground share to $36\%$ so that $\hat\tau_O$ falls at the valley.}
    \label{fig:gating}
\end{figure}

\Cref{fig:full-lvis} shows the failure mode on LVIS-92$^i$: over the full image the score mixture is effectively unimodal---one dominant background mode with a thin foreground tail---so Otsu cuts inside the background bulk (median $-0.27$, vs.\ $\tau^\ast$ at $+1.58$), costing $26.3$ mIoU.
FROST's forward--backward candidate gating~\cite{park2026frost}, which vetoes confidently identified background, changes the picture: restricted to the candidate region---the \emph{right part} of the score distribution---the mixture becomes roughly bimodal, exactly the regime classical separators assume, and Otsu lands at $+1.66$, nearly coinciding with $\tau^\ast$ ($+1.60$), at a cost of only $4.0$ mIoU (\Cref{fig:gated-lvis}).
The gate is thus what makes the separator viable; \Cref{sec:method} replaces it by a window read from the score itself and corrects the residual overshoot by shrinkage.

\section{Methodology}
\label{sec:method}

We present adaptive thresholding for KDE-based probabilistic models in four
steps: (1)~fixing the bandwidth (\Cref{sec:method:bandwidth}), (2)~augmenting the score field with a correspondence term
(\Cref{sec:method:score}), (3)~choosing the statistics window, the separator
and its correction (\Cref{sec:method:threshold}), and (4)~applying the
backward, i.e., query--support, correspondence vote as a veto
(\Cref{sec:method:veto}).

We pursue two goals. The first is reductive: strip
FROST~\cite{park2026frost} to the components that carry its performance,
leaving a pipeline that is nearly parameter-free. We drop support-side
augmentation, since mirroring a view is not new evidence; we remove bilateral
propagation; and we simplify its veto. The second is additive: replace the
fixed Bayes cut $\tau_0$ with a threshold estimated per episode from the
query's own score distribution.

\subsection{Bandwidth Selection}
\label{sec:method:bandwidth}

As noted in \Cref{sec:preliminaries:kde}, FROST~\cite{park2026frost} selects
the bandwidth per episode by maximizing the LOO margin
$M(\sigma)$ of \Cref{eq:loo_margin} over a small grid. The criterion is
unbounded in sharpness: as $\sigma \to 0$ the vMF kernel degenerates to
nearest-neighbor matching, $\ell_\sigma^{\setminus a}(a) \approx
\big(\cos_{\mathrm{nn}}^{+} - \cos_{\mathrm{nn}}^{-}\big)/\sigma$, so the
signed margin grows without limit and the maximizer is the smallest grid value
in essentially every episode. In effect, the procedure returns a constant.

A bounded LOO criterion selects a sharper bandwidth and performs worse; LOO is ill-posed on support patches, whose nearest neighbour is a near-duplicate.
We therefore state the constant directly, $\sigma = 0.05$, deleting the
selection stage and leaving the value unchanged. This keeps the comparison
controlled, as we then differ from FROST only in the components we change.

\subsection{Score Field}
\label{sec:method:score}
\label{sec:method:fwd}

We now augment FROST's score field with a support--query correspondence term. The density ratio $\ell_\sigma$ asks whether a query token is statistically
more compatible with the support foreground than with its background. It does
not ask whether that token is \emph{directly supported} by the annotated
foreground region. The distinction matters for frozen self-supervised features:
beyond organizing tokens by distributional structure, such representations
carry dense support--query semantic correspondences, which training-free
methods such as INSID3~\cite{cuttano2026insid3} exploit. The two sources of
evidence are complementary --- one a class-conditional comparison, the other a
matching term --- and we combine them.

Let $\operatorname{fwd}(q)$ measure how much of the support foreground's
attention lands on the query token $q$, computed on unwhitened features. For each anchor we take a softmax over the \emph{query} tokens and sum
the resulting mass at $q$:
\begin{equation}
  \log \sum_{a \in \mathcal{A}_{\mathrm{FG}}}
      \frac{\exp\!\big(\cos(q,a)/T\big)}
           {\sum_{q'}\exp\!\big(\cos(q',a)/T\big)}
    \;-\; \log\frac{N_{\mathrm{FG}}}{N_{\mathrm{Q}}}.
    \label{eq:fwd_calibrated}
\end{equation}
Normalizing over queries is what makes the term interpretable without a prior:
each foreground anchor distributes exactly one unit of attention mass across
the query image, so the total mass is $N_{\mathrm{FG}}$, and subtracting the
uniform value $\log(N_{\mathrm{FG}}/N_{\mathrm{Q}})$ fixes the chance level at
zero, giving $\mathbb{E}_q[\exp\operatorname{fwd}] = 1$ for any support set.
Hence $\operatorname{fwd}(q) > 0$ means that support foreground anchors select
$q$ more than chance would.

We combine the two signals additively,
\begin{equation}
  s(q) = \ell_\sigma(q) + \lambda \operatorname{fwd}(q),
  \label{eq:score}
\end{equation}
which acts as a tempered, likelihood-like correction of the density ratio.
Its influence recedes as the support grows: since $\operatorname{fwd}$ is an average over foreground anchors, accumulating anchors averages more attention maps over the same query image, flattening the term and narrowing its spread.
The density ratio behaves oppositely: its log-sum-exp terms act as soft maxima, so more anchors on both sides yield better-matched neighbors and a more extreme, better-separated ratio. The score is thus correspondence-assisted when evidence is scarce and density-dominated once it is plentiful, without any
schedule on $\lambda$.


\subsection{Adaptive Threshold}
\label{sec:method:threshold}

As established in \Cref{sec:preliminary}, a classical separator recovers the
cut well, but only on a restricted part of the score distribution. What follows
makes this operational (\Cref{fig:overview}): we choose the region the statistics are read from, then
the separator, then correct its bias.

\paragraph{Score-distribution window.}
A full-image histogram is dominated by background that is unambiguously
background, and that mass pulls every classical separator toward the bulk of
the distribution rather than toward the class boundary (\Cref{fig:full-lvis}).
\Cref{sec:preliminary} used FROST's candidate gate to show that \emph{some}
restricted region makes the separator viable; we observe that our score already supplies one, its own positive part,
\begin{equation}
  \mathcal{R} = \{\, q : s(q) > 0 \,\},
  \label{eq:window}
\end{equation}
the region on which the score itself votes foreground. $\mathcal{R}$ keeps the object together with the background the score is unsure about, and discards the background it is sure of --- precisely the mass that pulls a separator toward the bulk. What remains are the two populations it has to tell apart.

\paragraph{Separator.}
\label{sec:method:separator}

On the selected histogram we apply the minimum
method~\cite{prewitt1966valley}, in the iterated-smoothing form described by
Glasbey~\cite{glasbey1993analysis}: the histogram is smoothed repeatedly until
exactly two modes remain, and the cut $\hat{\tau}_{V}$ is placed at the minimum
between them; we call this the \emph{valley}. We prefer it to alternatives such
as Otsu's method~\cite{otsu1979otsu} because it locates the boundary between
the two modes irrespective of their relative mass, whereas Otsu's between-class
variance is mass-weighted and is dragged toward the larger mode --- the failure
of \Cref{fig:full-lvis}. Lastly, when no amount of smoothing yields a bimodal
histogram, we fall back to Otsu's cut.

\paragraph{Valley gate.}
$\mathcal{R}$ can nonetheless be uninformative for a given episode. We
therefore form both candidate histograms --- over $\mathcal{R}$ and over the
full query field --- smooth each as above,
and keep whichever yields the deeper valley, resolving ties in favor of the
full field. The choice is made per episode from the histograms alone.

\begin{figure*}[t]
    \centering
    \includegraphics[width=0.9\textwidth]{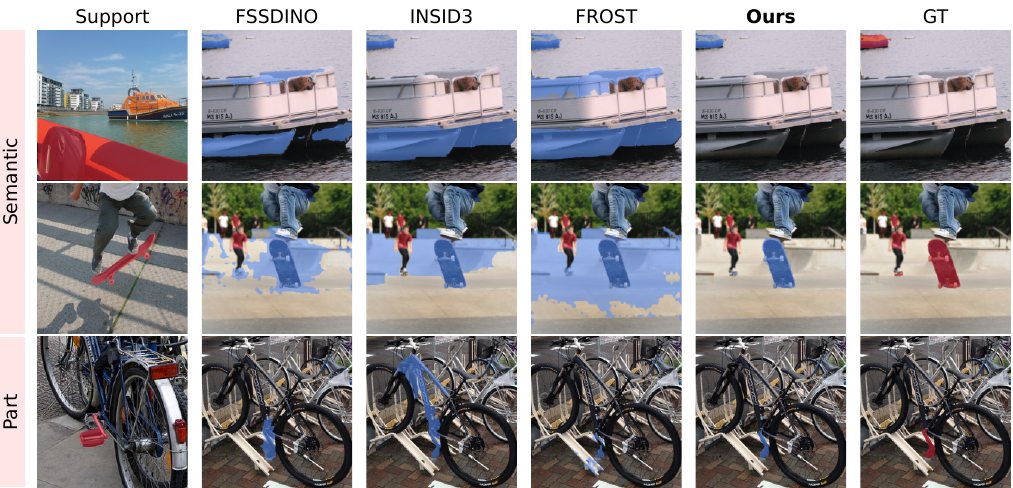}
    \caption{\textbf{Qualitative results.} 1-shot episodes on semantic and part benchmarks; ground truth (GT) in red, predictions in blue.}
    \label{fig:qualitative}
\end{figure*}

\paragraph{Shrinkage.}
\label{sec:method:shrinkage}
The background $\mathcal{R}$ retains is only the part the score was unsure
about, so its lower mode is thin and the valley lands slightly high, leaving
the cut biased towards the object. We pull it back multiplicatively, since the
score's scale varies from episode to episode and the valley drifts outward as
anchors accumulate --- an additive shift is invariant to neither. With
$\hat{\pi}$ the foreground fraction induced by Otsu's cut on the same
histogram, we set
\begin{equation}
  \hat{\tau} = \frac{\hat{\tau}_{V}}{1 + \hat{\pi}},
  \label{eq:shrinkage}
\end{equation}
moving the cut toward zero by the fraction $\hat{\pi}/(1+\hat{\pi})$.
On $\mathcal{R}$ the valley is positive by construction, so this is a pull toward the background; on the full-field fallback it may be negative, in which case the same rule pulls toward the object instead, a case we do not treat separately.
The shrinkage grows with $\hat{\pi}$ --- the more of the window is object, the deeper inside it the valley lies --- and vanishes as $\hat{\pi} \to 0$, leaving small objects untouched, the regime where pulling toward the background would be most damaging.
The cut is thereby only as good as $\hat{\pi}$, being driven entirely by Otsu's estimate of the foreground
fraction.

\subsection{Backward Correspondence Veto}
\label{sec:method:veto}

A complementary and largely independent check is available from the support
side: whether a query token is reciprocated by the support foreground. The
signal is exploited by INSID3~\cite{cuttano2026insid3} and adapted by
FROST~\cite{park2026frost}.

For each support view, we take the three support patches most similar to a given
query token; the view votes foreground if \emph{any} of them lies inside the support mask (a simplification \& relaxation to \cite{park2026frost}), and with several shots, the token passes when a majority of views
agree. We apply the result as a veto on the final segmentation,
\begin{equation}
  \hat{\mathcal{M}} = \{\, q : s(q) > \hat{\tau} \,\} \cap \mathcal{V},
  \label{eq:veto}
\end{equation}
where $\mathcal{V}$ is the set of tokens passing the vote. The veto filters the
thresholded mask and never enters the histogram that produced $\hat{\tau}$.
Because the vote is an agreement across views, each further view makes it
harder to pass: precision rises with the number of shots, recall does not. It
therefore contributes most where the correspondence term of
\Cref{sec:method:fwd} contributes least, and neither needs a schedule.

\section{Results}
\label{sec:results}
In the following, we will dive into our results, starting with the evaluation protocol, the baselines, and implementation details, followed by the comparison to the state of the art (\Cref{sec:results:main}) and an ablation of every component of the adaptive cut (\Cref{sec:results:ablation}).

\paragraph{Benchmark.} We follow the INSID3 protocol~\cite{cuttano2026insid3} and evaluate our adaptive thresholding rule on nine benchmarks.
For semantic segmentation --- segment every instance of a class, e.g., ``car'' ---
COCO-20$^i$~\cite{nguyen2019featurecoco} (80 classes), LVIS-92$^i$~\cite{liu2024matcher}
(920 long-tailed classes), ISIC2018~\cite{codella2018skin} (skin lesions),
Chest X-Ray~\cite{candemir2014lung}, iSAID-5$^i$~\cite{yao2022scale} (remote sensing, 15 classes)
and SUIM~\cite{islam2020semantic} (underwater, 8 classes). For part segmentation, e.g., ``car wheel'', PASCAL-Part~\cite{liu2024matcher} (56 parts, 15 classes) and PACO-Part~\cite{liu2024matcher} (303 parts, 75 classes).
For one-shot personalized segmentation, e.g., ``this car'', we use PerMIS~\cite{samuel2024where}, covering 16
categories.
In each episode, $k$ support images + mask and one query are sampled.
We report mIoU as in~\cite{cuttano2026insid3,park2026frost}: the IoU of the target class accumulated over all episodes of a fold, averaged over classes and folds.

\begin{table*}[t]
\centering
\small
\setlength{\tabcolsep}{4pt}
\scalebox{0.95}{
\begin{tabular}{l cccccccc c c}
\toprule
& \multicolumn{6}{c}{Semantic} & \multicolumn{2}{c}{Part} & Personalized & \\
\cmidrule(lr){2-7} \cmidrule(lr){8-9} \cmidrule(lr){10-10}
Method & LVIS-92$^i$ & COCO-20$^i$ & ISIC & SUIM & iSAID & X-Ray & PASCAL & PACO & PerMIS & Avg \\
\midrule
\rowcolor{gray!15}
\multicolumn{11}{c}{1-Shot} \\
\midrule
Matcher$^\dagger$ \cite{liu2024matcher} & 33.0 & 52.7 & 38.6 & 44.1 & 33.3 & 70.8 & 42.9 & 34.7 & 63.8 & 46.0 \\
GF-SAM$^\dagger$ \cite{zhang2024bridge} & 35.2 & 58.7 & 48.7 & 53.1 & 47.1 & 51.0 & 44.5 & 36.3 & 54.1 & 47.6 \\
\midrule
FSSDINO \cite{zakir2026fssdino} & 33.8 & 51.3 & 53.0 & 52.5 & 47.2 & 76.7 & 50.3 & 35.2 & 48.5 & 49.8 \\
INSID3 \cite{cuttano2026insid3} & \underline{41.5} & \underline{58.3} & 54.0 & 53.4 & \underline{51.7} & 78.6 & 49.1 & \underline{38.5} & \underline{67.0} & \underline{54.7} \\
FROST \cite{park2026frost} & 32.3 & 48.1 & \underline{54.4} & \underline{58.1} & 49.1 & \underline{84.8} & \underline{54.3} & 36.8 & 66.6 & 53.8 \\
\rowcolor{red!10}
Ours & \textbf{42.3} & \textbf{58.7} & \textbf{58.4} & \textbf{58.4} & \textbf{52.1} & \textbf{90.5} & \textbf{56.6} & \textbf{42.3} & \textbf{67.8} & \textbf{58.6} \\

\midrule
\rowcolor{gray!15}
\multicolumn{11}{c}{5-Shot} \\
\midrule
Matcher$^\dagger$ \cite{liu2024matcher} & 40.0 & 60.7 & 35.0 & 50.6 & 34.3 & 71.2 & 45.8 & 33.6 & -- & 46.4 \\
GF-SAM$^\dagger$ \cite{zhang2024bridge} & 44.2 & 66.8 & 55.2 & 58.1 & 52.4 & 52.9 & 51.2 & 41.9 & -- & 52.8 \\
\midrule
FSSDINO \cite{zakir2026fssdino} & 43.4 & 60.6 & 63.6 & 62.8 & 58.5 & 78.6 & 60.8 & 45.8 & -- & 59.2 \\
INSID3 \cite{cuttano2026insid3} & \underline{47.5} & \underline{64.3} & 63.9 & 63.1 & 60.1 & 79.4 & 58.4 & 47.0 & -- & 60.5 \\
FROST \cite{park2026frost} & \underline{47.5} & 61.8 & \textbf{66.7} & \textbf{67.2} & \underline{61.0} & \underline{86.4} & \underline{68.0} & \underline{51.3} & -- & \underline{63.7} \\
\rowcolor{red!10}
Ours & \textbf{52.2} & \textbf{67.3} & \underline{65.3} & \underline{66.8} & \textbf{61.9} & \textbf{91.8} & \textbf{68.4} & \textbf{54.3} & -- & \textbf{66.0} \\
\bottomrule
\end{tabular}
}
%
\caption{\textbf{Main results.} 1-shot \& 5-shot mIoU of training-free FSS methods; best in bold, second best underlined. PerMIS is one-shot only and evaluated here alone; all other experiments use the eight remaining benchmarks. $^\dagger$ results are taken over from INSID3~\cite{cuttano2026insid3} (not ranked; added for completeness) and denote training-free variants which received mask supervision during pre-training.}
\label{tab:results_no_permis}
\end{table*}

\paragraph{Baselines.}
FROST~\cite{park2026frost}, the method our pipeline is derived from, as well as INSID3~\cite{cuttano2026insid3} and
FSSDINO~\cite{zakir2026fssdino}.
All three are training-free FSS methods on frozen DINOv3 ViT-L/16 features, run from their public implementations on our episodes; Matcher and GF-SAM ($^\dagger$), which rely on SAM~\cite{kirillov2023sam}, are taken from~\cite{cuttano2026insid3}.

\paragraph{Implementation details.}
All images are resized to $1024\times 1024$~px, giving a $64\times64$ patch grid; the score is upsampled bilinearly to image size before it is thresholded.
All constants are fixed across datasets, domains, and tasks: $\sigma=0.05$, $T=0.1$, $\lambda=0.5$ (i.e., $\rho=\lambda\sigma/T=0.25$), $a=1$.
As $\ell_\sigma$ scales as $1/\sigma$ and $\operatorname{fwd}$ as $1/T$, $\lambda$ enters only through the dimensionless $\rho$; with $\sigma$ inherited from FROST, $T$ and $\rho$ were selected once by leave-one-dataset-out on our ablation split (fold~0, 250 episodes per dataset, eight benchmarks); the main table uses all folds and all episodes with a separate seed and thus different episodes.

\subsection{Main Results}
\label{sec:results:main}

\Cref{tab:results_no_permis} compares the six methods at one and five shots.
At one shot we lead on every benchmark, by $+4.8$~mIoU on average over FROST and $+3.9$ over INSID3.
The gains concentrate where the equal-prior assumption is furthest from the truth: on the object-centric COCO-20$^i$ and LVIS-92$^i$, where FROST trails INSID3, we gain $+10.6$ and $+10.0$ over it, and even on X-Ray, where the equal-prior regime roughly holds, $+5.7$.
At five shots the margin narrows to $+2.3$ and we cede two benchmarks, ISIC ($-1.4$) and SUIM ($-0.4$); \Cref{sec:results:ablation} shows both losses to be caused by the backward veto.
\Cref{fig:qualitative} shows one episode from LVIS-92$^i$ \& COCO-20$^i$ for semantic, and PACO-Part for part segmentation.
The baselines fail in different ways: FSSDINO's prototypes average the support and loses whatever departs from that mean; INSID3's partition, cut at a preset granularity, is coarser than the target and swallows it; and FROST's zero cut spills the mask into background that resembles the object. The adaptive cut, however, stops at the object.

\begin{figure}
    \centering
    \includegraphics[width=0.45\textwidth]{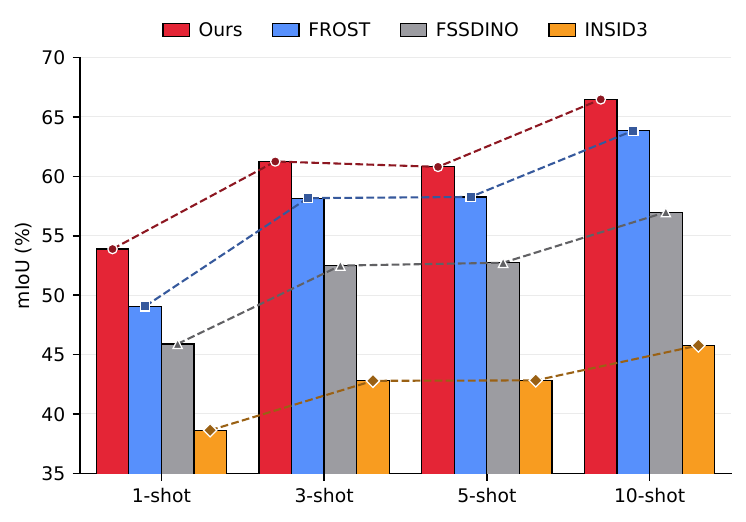}
    \caption{\textbf{Shot scaling.} $k$-shot mean mIoU over the eight benchmarks, employing a ViT-B/16 encoder. Our method leads at every $k$; the margin over FROST falls from $+4.8$ at one shot to $+2.6$ at ten, narrowing but not closing.}
    \label{fig:shot-scaling}
\end{figure}

\Cref{fig:shot-scaling} extends the comparison from one to ten shots. The
ordering of the four methods never changes, and our margin over FROST is
largest where support is scarcest, $+4.8$ at one shot, and settles at $+2.6$
by ten: it shrinks, as an adaptive threshold and a correspondence term should
buy the most under weak evidence, but it does not vanish.
For compute reasons, we use
ViT-B/16, which costs all methods roughly $5$~mIoU against
\Cref{tab:results_no_permis}, so only shape and ordering transfer --- with one instructive exception: INSID3 collapses on X-Ray at ViT-B/16 ($31.7$ against $79.0$ at ViT-L/16) which we attribute to its clustering granularity ($\tau=0.6$, cf.~\cite{cuttano2026insid3}). 
A fixed partition granularity is exactly the brittleness illustrated in \Cref{fig:teaser}, whereas a per-episode threshold has nothing to transfer.

\begin{table*}[t]
    \centering
    \scalebox{0.95}{%
    \begin{tabular}{lrrrr}
        \toprule
        & \multicolumn{2}{c}{1-shot} & \multicolumn{2}{c}{5-shot} \\
        \cmidrule(lr){2-3}\cmidrule(lr){4-5}
        Configuration & mIoU & $\Delta$ & mIoU & $\Delta$ \\
        \midrule
        \rowcolor{gray!15}
        FROST, full pipeline, our run~\cite{park2026frost} & 52.44 & -- & 61.65 & -- \\
        \midrule
        Base: fixed $\sigma$, $\tau_0$, no gate / flip / propagation & 48.23 & -- & 52.01 & -- \\
        \;+ correspondence term ($\tau_0$) & 50.17 & +1.94 & 53.17 & +1.16 \\
        \;+ adaptive threshold (valley gate) & 57.33 & +7.16 & 61.69 & +8.52 \\
        \;+ shrinkage & 58.20 & +0.87 & 62.74 & +1.06 \\
        \rowcolor{red!10}
        \;+ backward veto & \textbf{58.93} & +0.72 & \textbf{65.36} & +2.62 \\
        \midrule
        \rowcolor{gray!15}
        Oracle cut $\tau^*$ & 66.76 & -- & 71.34 & -- \\
        \bottomrule
    \end{tabular}%
    }
    \caption{\textbf{Ablation study.} Adding one component at a time; $\Delta$ is the gain over the row above, $\tau^*$ is the optimal cut.}
    \label{tab:ablation}
\end{table*}

\subsection{Ablation}
\label{sec:results:ablation}

\Cref{tab:ablation} builds the method up from FROST, one component at a time, in the order of \Cref{sec:method}.
The first row is the most instructive: stripped to the density ratio under the fixed cut $\tau_0$, FROST loses $4.2$~mIoU at one shot but $9.6$ at five. The base does not miss the object, it floods --- on LVIS-92$^i$ the predicted area is $5.9\times$ the object at one shot and $8.2\times$ at five --- since at $\sigma=0.05$ the ratio acts like nearest-neighbour matching, and every added support hands each background patch more anchors to match. FROST's candidate gate holds this flood back, and its precision rises with the number of views; the fixed cut itself degrades with evidence.
The remaining rows recover from this base. The correspondence term helps already
under the fixed cut ($+1.9$ / $+1.2$). The adaptive threshold is the largest
single step, $+7.2$ / $+8.5$: it exploits the better ranking more supports provide (the oracle rises from 66.8 to 71.3) instead of drowning in it.
Shrinkage ($+0.9$ / $+1.1$) and the backward veto ($+0.7$ /
$+2.6$) add the rest, and the veto's share keeps growing, to $+4.7$ at ten
shots: by five shots it already outweighs the correspondence term, which fades as anchors accumulate (see \Cref{sec:method:score}).
The oracle cut on the same field still sits $7.8$ / $6.0$ above us: the headroom lies not only in the ranking but still in the cut.

\begin{table}
    \centering
    \setlength{\tabcolsep}{4pt}
    \scalebox{0.95}{%
    \begin{tabular}{lccccc}
        \toprule
        Window & $|\mathcal{R}|/N_{\mathrm{Q}}$ & Otsu & Kittler & Valley & Median \\
        \midrule
        \rowcolor{gray!15}
        none & 1.00 & 53.9 & 55.7 & 58.0 & 56.2 \\
        \midrule
        FROST \cite{park2026frost} & 0.27 & 56.8 & 56.9 & 58.4 & 57.7 \\
        $\{\ell>0\}$ & 0.27 & 56.8 & 57.3 & \underline{58.8} & 57.8 \\
        $\{\mathrm{fwd}>0\}$ & 0.30 & 55.7 & 56.3 & 58.3 & 57.1 \\
        \rowcolor{red!10}
        $\{s>0\}$ & 0.25 & 57.1 & 57.4 & \textbf{58.9} & 58.1 \\
        \bottomrule
    \end{tabular}%
    }
    \caption{\textbf{Window $\times$ separator.} 1-shot, mean mIoU over the eight benchmarks. $|\mathcal{R}|/N_{\mathrm{Q}}$ is the window's mean image coverage.}
    \label{tab:window}
\end{table}

\paragraph{Which window, which separator?}
\Cref{tab:window} crosses five statistics windows with four classical
separators, everything else as deployed. Two things stand out: the valley is the best separator in every window, by $2$--$4$~mIoU over Otsu, and it depends on the window far less: restricting the histogram from the full field to $\{s>0\}$ gains Otsu $+3.2$ but the valley only $+0.9$, because Otsu is dragged toward the background bulk (see \Cref{fig:full-lvis}) and the valley is not.
Among windows, $\{s>0\}$ matches or beats every alternative, including FROST's candidate region ($+0.5$) and either of its two terms alone, and it is narrow, covering a quarter of the image. A bimodal histogram is found in $99.8$\,\% of episodes, so the Otsu fallback is essentially never taken; the valley gate, which keeps the window in $35$\,\% of the episodes and reads the full field otherwise, is what protects the rule when the window $\mathcal{R}$ is uninformative.

\paragraph{Does the shrinkage matter?}
When sweeping $a$ in
$\hat\tau=\hat\tau_V/(1+a\hat\pi)$ over $\{0,0.5,1,2,4\}$, unit strength sits on a plateau: $a$ peaks on $[1,2]$ at every shot count we tested, and removing the correction ($a=0$) costs $0.5$ / $1.0$. The form matters more than the strength: an additive shift by $\hat\pi$ loses $0.5$ / $0.3$ against the multiplicative one, because the
relative correction stays at $0.21$--$0.22$ from one to ten shots while the
valley itself drifts outward with more anchors; a ratio follows that drift, an offset does not.

\Cref{tab:ablation}'s order also hides an interaction.
Since $\mathcal{R}$ carries no full background mode (\Cref{sec:method:shrinkage}), the raw valley sits inside the object, and two devices pull it back: the shrinkage, and the valley gate that falls back to the full-field histogram when its valley is deeper.
They are substitutes: each rescues the positives-only valley on its own (worth $6.0$ and $8.3$~mIoU at one shot when the other is absent) and adds only $0.5$ / $2.8$ given the other.
The valley gate's value is not uniform, either: at five shots, without the veto, it gains $6.7$ on ISIC and $8.9$ on SUIM but costs $7.2$ on LVIS-92$^i$ and $6.0$ on PACO-Part.
It trades on small-object data for robustness where the positive window is uninformative.

\paragraph{Does the backward veto help, and when?}
The veto is the complement of the correspondence term: $+0.7$, $+1.7$, $+2.6$
and $+4.7$~mIoU at one, two, five and ten shots, and it is what makes the
five-shot margin over FROST.
The mechanism is the vote's own precision, which rises from $0.61$ at one shot to $0.77$ at ten while its recall stays at $0.95$--$0.99$ (medians over episodes): agreement across views removes false
positives without removing the object.
FROST's full gate in the same position is slightly worse ($0.5$ / $0.3$).
The veto is also where the two five-shot losses of \Cref{tab:results_no_permis} come from: on ISIC and SUIM it costs $5.7$ and $2.8$, removing foreground that the support views fail to reciprocate. 

\section{Conclusion}
In this paper, we set out to replace the fixed Bayes cut of KDE-based few-shot segmentation with a threshold read from the query score field itself.
Built on the \textit{minimum} method~\cite{prewitt1966valley} and free of any threshold constant, our rule leads on all nine benchmarks at one shot and, on the eight that provide more, from one to ten shots, with the adaptive cut itself the largest step of the ablation.
Two limitations remain: (1) a density ratio needs both classes in the support, so a mask covering the whole image, or an object that is a sliver of it, leaves one density undefined or poorly estimated, though such cases grow rarer with more shots; and (2) the score is purely visual, so a target semantically but not visually related to its support is out of reach without semantic reasoning over the support.
Likewise, three headrooms remain: (1) the oracle cut still sits $7.8$~mIoU above ours on the same field, so the cut is not yet solved; (2) the oracle itself is bounded by the ranking, which per-episode feature adaptation~\cite{herzog2024adapt,nie2026mpa} or multimodal cues~\cite{chen2025cmp,catalano2026mars} could lift; and (3) both are read from a $64\times64$ grid, which adaptive patching~\cite{rohan2026faster} could refine --- all orthogonal to our contribution and promising to compound with it.

{
    \small
    \bibliographystyle{ieeenat_fullname}
    \bibliography{main}
}

\clearpage
\appendix
\twocolumn[{%
  \centering
  {\Large\bfseries Supplementary Material\par}
  \vspace{6pt}
  {\large Where to Cut? Adaptive Thresholds for Density-Ratio Few-Shot Segmentation\par}
  \vspace{8pt}
  \rule{0.6\textwidth}{0.4pt}\par
  \vspace{10pt}
}]
\suppressfloats[t]
\setcounter{section}{0}
\renewcommand{\thesection}{\Alph{section}}
\setcounter{table}{0}
\renewcommand{\thetable}{S\arabic{table}}
\setcounter{figure}{0}
\renewcommand{\thefigure}{S\arabic{figure}}
\setcounter{equation}{0}
\renewcommand{\theequation}{S\arabic{equation}}
\crefname{appendix}{Appendix}{Appendices}
\Crefname{appendix}{Appendix}{Appendices}
This supplement backs the constants and the analysis of the main paper with
the tables behind them: \Cref{sec:supp:perds} breaks the ablation of
\Cref{tab:ablation} down by benchmark, \Cref{sec:supp:qualitative} extends
\Cref{fig:qualitative}, 
\Cref{sec:supp:bandwidth} backs the bandwidth of \Cref{sec:method:bandwidth},
\Cref{sec:supp:score} the score field of \Cref{sec:method:score},
\Cref{sec:supp:threshold} the adaptive threshold of \Cref{sec:method:threshold},
and \Cref{sec:supp:veto} the backward veto of \Cref{sec:method:veto}. Unless stated otherwise, every experiment uses the ablation protocol of the main paper and reports the 8-benchmark mean mIoU; two values separated by a slash are one / five shots.

\section{Per-Dataset Ablation}
\label{sec:supp:perds}

\Cref{tab:supp:perds} breaks the ablation of \Cref{tab:ablation} down by benchmark.
Three observations the mean hides:
First, the KDE base under the
fixed cut $\tau_0$ collapses most on the object-centric and part benchmarks: on
LVIS-92$^i$ it loses $6.8$ / $22.2$~mIoU against FROST's full pipeline, on
PASCAL-Part $10.0$ / $17.5$, while on ISIC and SUIM
it stays within about one point of it, and on X-Ray within about five. Second, the adaptive threshold is the
largest step on every benchmark at one shot except SUIM, where it costs $3.0$
under the fixed cut and the shrinkage recovers $2.3$ of that; at five shots, it
is the largest step on six of eight, the exceptions again being ISIC ($+0.6$)
and SUIM ($+1.1$), where the shrinkage contributes more. Third, the backward veto
is what lifts LVIS-92$^i$ ($+4.4$ / $+10.5$) and, at five shots, COCO-20$^i$,
PASCAL-Part and PACO-Part ($+5.5$, $+5.2$, $+5.7$), and it is the sole source of the two five-shot losses reported in the
main paper: ISIC $68.9 \to 63.2$ and SUIM $66.0 \to 63.2$. The oracle cut on
the same field leaves the most headroom on ISIC and LVIS-92$^i$ at one shot
($11.6$, $11.5$) and on LVIS-92$^i$ at five ($9.0$), and the least on X-Ray
($0.9$ / $0.7$).

\begin{table*}[t]
  \centering
  \setlength{\tabcolsep}{4pt}
  \scalebox{0.95}{%
  \begin{tabular}{lccccccccc}
    \toprule
    Configuration & LVIS-92$^i$ & COCO-20$^i$ & ISIC & SUIM & iSAID & X-Ray & PASCAL & PACO & Avg \\
    \midrule
    \rowcolor{gray!15}
    \multicolumn{10}{c}{1-Shot} \\
    \midrule
    FROST, full pipeline, our run~\cite{park2026frost} & 31.6 & 48.1 & 57.9 & 56.2 & 57.9 & 85.4 & 47.5 & 34.9 & 52.4 \\
    \midrule
    Base: fixed $\sigma$, $\tau_0$, no gate / flip / propagation & 24.8 & 43.9 & 57.3 & 56.5 & 55.9 & 81.0 & 37.5 & 29.0 & 48.2 \\
    \;+ correspondence term ($\tau_0$) & 25.5 & 45.8 & 57.6 & \textbf{58.2} & 58.4 & 82.7 & 40.8 & 32.2 & 50.2 \\
    \;+ adaptive threshold (valley gate) & 38.9 & 56.8 & \underline{61.3} & 55.2 & \underline{62.0} & \underline{89.8} & 52.9 & 41.7 & 57.3 \\
    \;+ shrinkage & \underline{39.3} & \underline{58.0} & \textbf{61.9} & \underline{57.5} & \textbf{62.7} & \textbf{90.9} & \underline{53.2} & \textbf{42.1} & \underline{58.2} \\
    \rowcolor{red!10}
    \;+ backward veto & \textbf{43.7} & \textbf{59.0} & 60.7 & 57.3 & \textbf{62.7} & \textbf{90.9} & \textbf{55.4} & \underline{41.8} & \textbf{58.9} \\
    \midrule
    \rowcolor{gray!15}
    \textit{Oracle cut $\tau^\ast$} & 55.2 & 69.7 & 72.3 & 63.3 & 70.9 & 91.8 & 61.3 & 49.6 & 66.8 \\
    \midrule
    \rowcolor{gray!15}
    \multicolumn{10}{c}{5-Shot} \\
    \midrule
    FROST, full pipeline, our run~\cite{park2026frost} & 42.6 & 62.9 & 66.4 & 61.7 & 69.1 & 86.3 & 60.0 & 44.3 & 61.7 \\
    \midrule
    Base: fixed $\sigma$, $\tau_0$, no gate / flip / propagation & 20.4 & 51.9 & \underline{67.5} & 61.5 & 63.4 & 81.1 & 42.5 & 27.7 & 52.0 \\
    \;+ correspondence term ($\tau_0$) & 20.7 & 53.9 & 66.2 & 62.9 & 64.9 & 82.5 & 43.9 & 30.5 & 53.2 \\
    \;+ adaptive threshold (valley gate) & 40.8 & 60.7 & 66.8 & \underline{64.0} & 70.2 & \underline{90.5} & \underline{59.0} & 41.4 & 61.7 \\
    \;+ shrinkage & \underline{41.2} & \underline{61.9} & \textbf{68.9} & \textbf{66.0} & \underline{71.9} & \textbf{91.9} & 58.4 & \underline{41.7} & \underline{62.7} \\
    \rowcolor{red!10}
    \;+ backward veto & \textbf{51.7} & \textbf{67.4} & 63.2 & 63.2 & \textbf{74.5} & \textbf{91.9} & \textbf{63.6} & \textbf{47.4} & \textbf{65.4} \\
    \midrule
    \rowcolor{gray!15}
    \textit{Oracle cut $\tau^\ast$} & 60.7 & 74.8 & 68.9 & 69.6 & 80.7 & 92.6 & 69.1 & 54.3 & 71.3 \\
    \bottomrule
  \end{tabular}}
  \caption{\textbf{Per-dataset ablation.} \Cref{tab:ablation} by benchmark (without PerMIS).
  Each row adds one component to the row above; the final row of each block is our deployed configuration; best per column among the configurations in bold, second underlined; the oracle cut is the best per-episode threshold on the same score field.}
  \label{tab:supp:perds}
\end{table*}

\begin{figure}[t]
  \centering
  \includegraphics[width=0.6\linewidth]{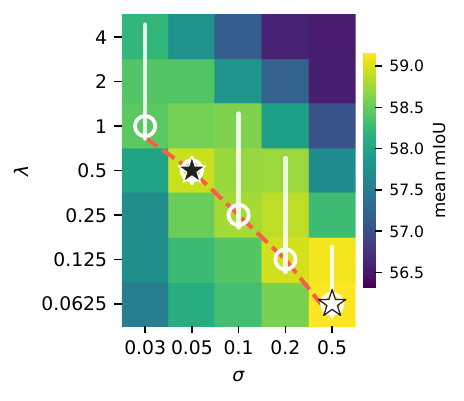}
  \caption{\textbf{Scale law along $\sigma$.} $\lambda^\star=\rho T/\sigma$ at the deployed configuration, $T=0.1$. Circles: each column's measured optimum; bars: the $\lambda$ within $0.3$~mIoU of it; dashed: the prediction with $\rho=0.25$; white star: the joint leave-one-dataset-out selection; black star: our operating point where the two differ.}
  \label{fig:ridge}
\end{figure}

\begin{figure}[b]
  \centering
  \includegraphics[width=0.6\linewidth]{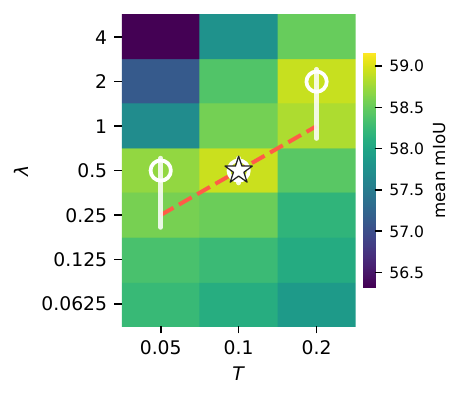}
  \caption{\textbf{Scale law along $T$.} Same conventions as \Cref{fig:ridge}, at $\sigma=0.05$; the joint selection coincides with our operating point.}
  \label{fig:ridge-temp}
\end{figure}

\section{Extended Qualitative Results}
\label{sec:supp:qualitative}

\begin{figure*}
  \centering
  \includegraphics[width=\textwidth]{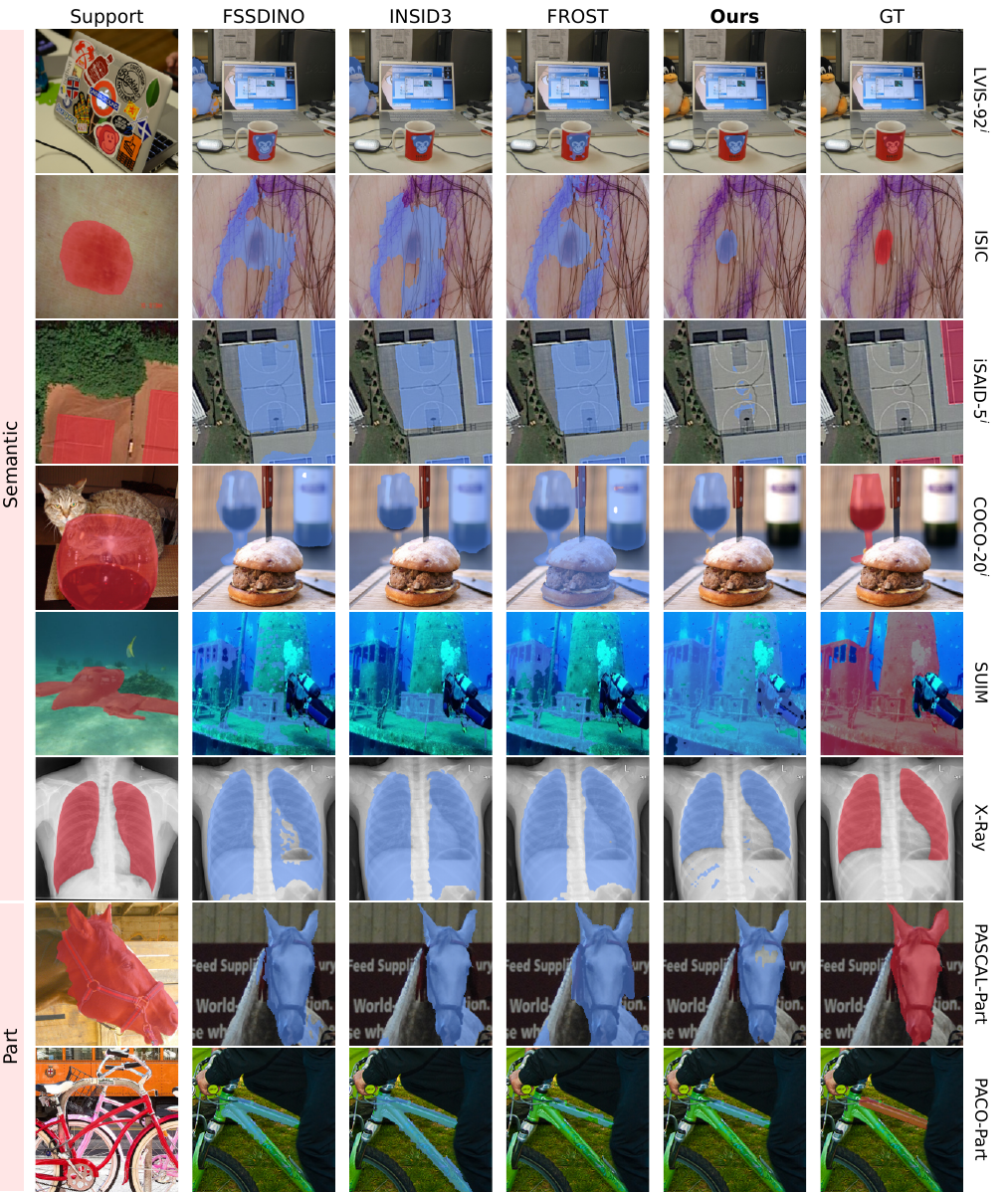}
  \caption{\textbf{Extended qualitative results.} One 1-shot episode per benchmark, semantic (top) and part (bottom); support annotation and ground truth (GT) in red, predictions in blue; the query tiles of a row share one square window centered on the object.}
  \label{fig:supp:qualitative}
\end{figure*}

\Cref{fig:supp:qualitative} extends \Cref{fig:qualitative} of the main paper to one episode
per benchmark. The pattern of \Cref{fig:qualitative} repeats across domains: FSSDINO's prototype loses what departs from the support mean, INSID3's
partition is either coarser than the target or merges it into a neighbor, and
FROST's fixed zero spills into the background, most
visibly on the small-object benchmarks (LVIS-92$^i$, iSAID-5$^i$, PACO-Part).

\section{Bandwidth}
\label{sec:supp:bandwidth}

\paragraph{How to choose the bandwidth?}
\Cref{tab:bandwidth} runs FROST's LOO margin criterion at the deployed
configuration, on FROST's grid and on that grid extended below its floor. It
selects the smallest value in $100\,\%$ of episodes on both --- the criterion
is unbounded in sharpness (\Cref{sec:method:bandwidth}) and returns whatever
floor it is given; on the published grid, this coincides with our fixed
$\sigma=0.05$, on the extended grid it costs $6.3$~mIoU. A bounded, proper
criterion (the class-balanced Bernoulli log-likelihood of the LOO
predictions) does select interior values, but prefers $\sigma=0.01$ and loses
$2.4$: sharpness that wins on the support does not transfer to the query.
Refining the criterion does not recover the gap: breaking ties toward the
smoother end (\emph{critical}) still lands on $\sigma=0.01$ and loses $2.8$,
and holding out each anchor's $3\times3$ neighbourhood --- so that the
near-duplicate neighbours of \Cref{sec:method:bandwidth} no longer vote --- moves the selection
only to $0.02$ and loses $2.2$; the $m$-standard-error rules, which take the largest $\sigma$
within $m$ standard errors of the maximum and so prefer the smoother end of the LOO curve, recover $0.4$ at $m=1$ and $1.1$ at
$m=3$ --- where the rule is so permissive that it barely constitutes a
criterion --- and still lose $2.0$ and $1.3$. Every variant selects a bandwidth in
$[0.01,0.02]$ while the ridge of \Cref{fig:ridge} is flat from $0.03$ to
$0.5$; the per-episode LOO evidence simply points the wrong way. We
therefore fix $\sigma$ at FROST's published value, which \Cref{fig:ridge}
shows to be one point on a nearly flat ridge.
Selecting constants on COCO alone, as INSID3 does, would favour COCO; apart from $\sigma$, inherited from FROST, our constants $T$, $\rho$ and $a$ are selected leave-one-dataset-out.

\section{Score Field}
\label{sec:supp:score}

\paragraph{One scale law.}
The two terms of the score $s = \ell_\sigma + \lambda\operatorname{fwd}$
(\cref{eq:score}) live on different scales: $\ell_\sigma$ is a
difference of log-sum-exp terms in $\cos(\cdot)/\sigma$, so
$\operatorname{std}(\ell_\sigma) \propto 1/\sigma$, while $\operatorname{fwd}$
is a softmax at temperature $T$, for which we observe
$\operatorname{std}(\operatorname{fwd}) \propto 1/T$. Commensurability
therefore requires
\begin{equation}
  \lambda^{\star} = \rho\,\frac{T}{\sigma},
  \label{eq:scalelaw}
\end{equation}
where $\rho = \lambda\sigma/T$ is a pure number --- the combination in which the
two scales cancel, and so the one quantity that stays fixed as $\sigma$, $T$, and
$\lambda$ move together. $\lambda$ is therefore not independently free. With
$\sigma = 0.05$ inherited from FROST~\cite{park2026frost}, what remains to
select is $T$ together with $\rho$, which leave-one-dataset-out sets to
$(0.1,\,0.25)$ --- interior on both axes, giving $\lambda = 0.5$.

\paragraph{Are $\sigma$, $T$ and $\lambda$ independent?}
\Cref{fig:ridge,fig:ridge-temp} sweep $\lambda$ against $\sigma$ at fixed $T$
and against $T$ at fixed $\sigma$. Along $\sigma$ the measured optimum lands on the grid
point nearest the prediction of \Cref{eq:scalelaw} with $\rho = 0.25$ at all five
bandwidths ($\lambda = 1, 0.5, 0.25, 0.125, 0.0625$ for
$\sigma = 0.03, 0.05, 0.1, 0.2, 0.5$); along $T$ the prediction lies within
$0.13$~mIoU of the optimum at both off-center temperatures. The ridge
$\sigma\lambda = \rho T$ is therefore an artifact of matching two scales, not a
property of the data --- and it is nearly flat: its five column optima span
$0.7$~mIoU ($58.4$--$59.1$). When performing leave-one-dataset-out over the whole
$\sigma$--$\lambda$ grid, it picks its large-bandwidth end ($\sigma = 0.5$,
$\lambda = 0.0625$ in six of eight folds, $\lambda = 0.125$ in the other two;
white star), $0.2$~mIoU above our operating point (black star) --- within the
protocol's resolution. Only $\rho$ is identified; the position along the ridge
is free, and we take it from FROST's published bandwidth to keep the
comparison controlled. Over $T$ the joint selection coincides with our
operating point.

\begin{figure}
  \centering
  \begin{subfigure}[t]{0.49\linewidth}
    \centering
    \includegraphics[width=\linewidth]{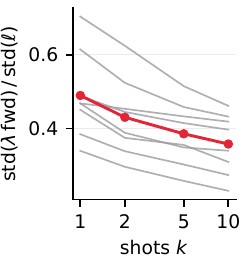}
    \caption{Correspondence term}
    \label{fig:k-fwd}
  \end{subfigure}
  \hfill
  \begin{subfigure}[t]{0.49\linewidth}
    \centering
    \includegraphics[width=\linewidth]{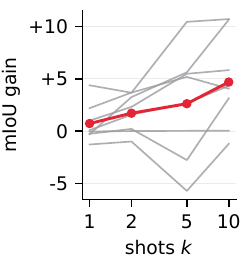}
    \caption{Backward veto}
    \label{fig:k-veto}
  \end{subfigure}
  \caption{\textbf{Shot behaviour.} (a)~Relative weight of the correspondence term, $\mathrm{std}(\lambda\,\mathrm{fwd})/\mathrm{std}(\ell)$, with $\lambda$ fixed; (b)~mIoU the backward vote adds over no veto. Per benchmark in grey, average in red. As the support grows, the correspondence term recedes while the veto's gain rises; neither $\lambda$ nor the vote rule changes with $k$.}
  \label{fig:k}
\end{figure}

\paragraph{Does the correspondence term help, and when?}
Under the fixed cut, it is worth $+1.9$ / $+1.2$ (\Cref{tab:ablation}).
Under the adaptive rule its contribution is isolated by keeping $\{s>0\}$ as
the window but thresholding the bare density ratio: $+1.3$ at one shot, a tie
($+0.2$) at five. The term cannot be demoted to choosing where to look --- its
value is in the ranking --- and it fades as the support grows, as \Cref{sec:method:score} predicts: \Cref{fig:k-fwd} shows the spread of
$\lambda\,\mathrm{fwd}$ relative to $\ell$ falling from $0.49$ at one shot to
$0.36$ at ten on average, with $\lambda$ fixed.

\paragraph{Degenerate supports.}
A support mask covering the whole grid leaves $\mathcal{A}_{\mathrm{BG}}$ empty
and the density ratio undefined. Since $\operatorname{fwd}$ needs no
background anchors, we fall back to it alone:
\begin{equation}
  s(q) =
  \begin{cases}
    \ell_\sigma(q) + \lambda \operatorname{fwd}(q), & N_{\mathrm{BG}} > 0,\\[2pt]
    \operatorname{fwd}(q),                          & N_{\mathrm{BG}} = 0.
  \end{cases}
  \label{eq:score_cases}
\end{equation}
This branch keeps a window of its own: by \Cref{eq:fwd_calibrated}
$\exp\operatorname{fwd}$ has unit mean over the query, so
$\operatorname{fwd}(q) > 0$ marks the tokens the support foreground selects
above chance --- a window in the spirit of positive--unlabelled
learning~\cite{elkan2008learning}, taken without a class prior --- and the
adaptive cut of \Cref{sec:method:threshold} is read from it as usual; the
veto is vacuous, as every support patch lies inside the mask. The case
occurred in 44 of 60{,}581 one-shot episodes of the full benchmark (31 iSAID,
11 LVIS, 1 COCO, 1 PACO) and never at five shots. Returning all-background
instead would score zero on them; the fallback recovers a mean IoU of $43.5$ on
those episodes ($97.4$ at best), worth $0.01$--$0.02$~mIoU on the benchmark
mean --- a correctness detail.

\begin{table}
  \centering
  \scalebox{0.95}{%
  \begin{tabular}{lrlr}
    \toprule
    Criterion & mIoU & mode $\sigma$ & picked \\
    \midrule
    \rowcolor{red!10}
    Fixed $\sigma=0.05$ (ours) & $\mathbf{58.9}$ & -- & -- \\
    \midrule
    LOO margin, published grid & $58.9$ & $0.05$ & $100\%$ \\
    LOO margin, extended grid & $52.7$ & $0.005$ & $100\%$ \\
    \midrule
    Bernoulli LOO & $56.5$ & $0.01$ & $40\%$ \\
    \;critical (largest tie) & $56.1$ & $0.01$ & $37\%$ \\
    \;spatial $3\times3$ holdout & $56.7$ & $0.02$ & $38\%$ \\
    \;one-standard-error rule & $56.9$ & $0.01$ & $32\%$ \\
    \;three-standard-error rule & $57.6$ & $0.02$ & $27\%$ \\
    \bottomrule
  \end{tabular}}
  \caption{\textbf{Bandwidth criteria.} Per-episode rules at one shot; ``picked'' is the share of episodes selecting the mode $\sigma$. FROST's grid is $\{0.05,0.1,0.2,0.5,1\}$; the extended grid (all Bernoulli rows) adds
 $\{0.005,0.01,0.02,0.03,0.04\}$.}
  \label{tab:bandwidth}
\end{table}

\begin{table}
  \centering
  \setlength{\tabcolsep}{4pt}
  \scalebox{0.95}{%
  \begin{tabular}{lccccc}
    \toprule
    Window & $|\mathcal{R}|/N_{\mathrm{Q}}$ & Otsu & Kittler & Valley & Median \\
    \midrule
    \rowcolor{gray!15}
    none & 1.00 & 62.0 & 63.1 & 65.3 & 63.6 \\
    \midrule
    FROST~\cite{park2026frost} & 0.23 & \textbf{63.9} & \underline{63.7} & 65.2 & \textbf{64.5} \\
    $\{\ell>0\}$ & 0.27 & 63.6 & \underline{63.7} & \textbf{65.5} & \underline{64.4} \\
    $\{\mathrm{fwd}>0\}$ & 0.31 & 63.0 & 63.4 & 65.1 & 63.9 \\
    $\{s>0\}$ dilated & 0.30 & 63.6 & 63.6 & \underline{65.4} & 64.3 \\
    \rowcolor{red!10}
    $\{s>0\}$ (ours) & 0.26 & \underline{63.8} & \textbf{63.9} & \underline{65.4} & \textbf{64.5} \\
    \bottomrule
  \end{tabular}}
  \caption{\textbf{Threshold ablation at five shots.} Statistics window $\times$ separator, everything else as deployed; best per column in bold, second underlined. $|\mathcal{R}|/N_{\mathrm{Q}}$ is the window's mean coverage of the query; ``dilated'' grows $\{s>0\}$ by FROST's support-adaptive radius $r=\operatorname{round}(4\pi_s)$; the oracle cut on the same field is $71.3$.}
  \label{tab:supp:window5}
\end{table}

\section{Adaptive Threshold}
\label{sec:supp:threshold}

\paragraph{Window and separator at five shots.}
\Cref{tab:supp:window5} repeats \Cref{tab:window} at five shots. The
ordering of separators is unchanged --- the valley leads every window by
$1.3$--$3.2$~mIoU over Otsu --- and the windows are closer together than at
one shot: $\{s>0\}$, $\{\ell>0\}$, FROST's candidate region and a dilated
$\{s>0\}$ lie within $0.25$ of each other under the valley, with $\{\ell>0\}$
nominally ahead of the deployed window by $0.1$. With five supports the
density ratio alone already separates the two populations, so the choice of
window matters less; we keep $\{s>0\}$ for its one-shot advantage. The
dilated window grows $\{s>0\}$ by FROST's support-adaptive radius,
$r=\operatorname{round}(4\,\pi_s)$ patches with $\pi_s$ the support foreground
fraction, and adds nothing at either shot count: the useful window is narrow.

\paragraph{Fallback and valley gate.}
The separator smooths the histogram with a $3$-tap kernel until exactly two
modes remain and cuts at the minimum between them. If the histogram is
unimodal to begin with, no amount of smoothing creates a second mode and we
fall back to Otsu's cut; this happens in $0.2\,\%$ of episodes at one shot
and $0.1\,\%$ at five. The valley gate then decides which histogram the cut is
read from: both candidates --- over $\{s>0\}$ and over the full field --- are
smoothed as above, and the one whose valley is deeper (the larger drop from
the lower of the two peaks to the minimum, relative to that peak) is kept,
ties resolving to the full field. The gate keeps the window in $35\,\%$ of
the episodes at one shot and $34.5\,\%$ at five.

\begin{table}[h]
  \centering
  \scalebox{0.95}{%
  \begin{tabular}{cccccc}
    \toprule
    \diagbox[width=2.4em,height=2.4em]{$k$}{$a$} & $0$ & $0.5$ & $1$ & $2$ & $4$ \\
    \midrule
    \rowcolor{gray!15}
    1   & $58.4$ & $\underline{58.8}$ & $\mathbf{58.9}$ & $58.7$ & $57.9$ \\
    2   & $61.9$ & $\underline{62.4}$ & $\mathbf{62.5}$ & $\underline{62.4}$ & $61.8$ \\
    \rowcolor{red!10}
    5   & $64.4$ & $\underline{65.1}$ & $\mathbf{65.4}$ & $\mathbf{65.4}$ & $65.0$ \\
    10  & $68.9$ & $69.6$ & $\mathbf{69.8}$ & $\underline{69.7}$ & $69.3$ \\
    \bottomrule
  \end{tabular}}
  \caption{\textbf{Shrinkage strength.} Best per row in bold, second underlined (the deployed $a=1$ is the middle column). The optimum is a plateau on $[1,2]$ at every shot count; both ends lose.}
  \label{tab:supp:shrinkage-sweep}
\end{table}

\paragraph{Shrinkage strength.}
\Cref{sec:method:shrinkage} corrects the valley multiplicatively at unit
strength, $\hat\tau=\hat\tau_V/(1+\hat\pi)$. \Cref{tab:supp:shrinkage-sweep}
sweeps the strength $a$ in $\hat\tau=\hat\tau_V/(1+a\hat\pi)$ at four shot
counts. The response is flat over $[1,2]$ and falls away at both ends:
removing the correction ($a=0$) costs $0.5$ / $1.0$, over-correcting ($a=4$)
costs $1.0$ / $0.4$. The unit value lies on the plateau at every shot count,
so it is adopted without being tuned per setting.

\paragraph{Shrinkage form.}
An additive shift by $\hat\pi$, $\hat\tau=\hat\tau_V-\hat\pi$, loses $0.5$ /
$0.3$ against the multiplicative form, for the reason given in
\Cref{sec:method:shrinkage}: the score's scale
varies from episode to episode, so a fixed offset means different things in
different episodes. $\hat\pi$, by contrast, is a property of the object rather
than of the support size, so the relative shrinkage $\hat\pi/(1+\hat\pi)$ stays
at $0.21$--$0.22$ from one to ten shots, while the valley itself drifts outward
as anchors accumulate, from a mean of $1.16$ at one shot to $1.52$ at ten. A
multiplicative correction follows that drift automatically.

\begin{table}[h]
    \centering
    \scalebox{0.95}{%
    \begin{tabular}{lrrrr}
    \toprule
    & \multicolumn{2}{c}{1-shot} & \multicolumn{2}{c}{5-shot} \\
    \cmidrule(lr){2-3}\cmidrule(lr){4-5}
    Shrinkage & gate off & gate on & gate off & gate on \\
    \midrule
    \rowcolor{gray!15}
    none ($a=0$) & 50.1 & 58.4 & 58.4 & 64.4 \\
    \rowcolor{red!10}
    unit ($a=1$) & 56.1 & \textbf{58.9} & 63.6 & \textbf{65.4} \\
    \bottomrule
    \end{tabular}}
    \caption{\textbf{Valley gate $\times$ shrinkage.} The two are substitutes: each rescues the positives-only valley on its own, and they overlap given the other; the deployed cell is bold.}
    \label{tab:gate-x-shrink}
\end{table}

\paragraph{Valley gate $\times$ shrinkage.}
\Cref{tab:gate-x-shrink} is the $2\times2$ behind the interaction discussed in
\Cref{sec:results:ablation}. Without the valley gate, shrinkage is worth $6.0$ /
$5.2$; with it, $0.5$ / $1.0$. The gate is worth $8.3$ / $6.0$ without
shrinkage and $2.8$ / $1.8$ with it. Each rescues the positives-only valley on
its own, so each is worth far less given the other.

\begin{table}[h]
  \centering
  \scalebox{0.95}{%
  \begin{tabular}{lcc}
    \toprule
    Veto & 1-shot & 5-shot \\
    \midrule
    \rowcolor{gray!15}
    FROST's full gate & \underline{50.3} & \textbf{56.5} \\
    \;$-$ forward test & \underline{50.3} & \textbf{56.5} \\
    \;$-$ dilation & 49.6 & 54.3 \\
    \rowcolor{red!10}
    \;$-$ strict vote (ours) & \textbf{50.8} & \underline{56.4} \\
    \bottomrule
  \end{tabular}}
  \caption{\textbf{Dissecting FROST's gate.} FROST's candidate gate used as the veto of our pipeline; each row removes one more component than the row above, the last relaxing the 2-of-3 per-view vote to any-of-3. Mean over the four benchmarks the gate affects most (LVIS-92$^i$, PACO-Part, ISIC, SUIM). Removing the forward test changes no episode; the dilation and the strict vote are opposing mechanisms that nearly cancel.}
  \label{tab:supp:veto-parts}
\end{table}

\section{Backward Veto}
\label{sec:supp:veto}

FROST's candidate gate bundles three mechanisms: a per-view vote (a query
token passes a view if at least two of its three nearest support patches lie
inside the mask, and passes overall if a majority of views agree), a forward
prototype test (positive cosine to the mean foreground feature), and a
spatial dilation of the result by $r=\operatorname{round}(4\,\pi_s)$ patches.
\Cref{tab:supp:veto-parts} removes them one at a time in our pipeline, on the four benchmarks the gate affects most (LVIS-92$^i$, PACO-Part, ISIC, SUIM), so its means are not comparable to the 8-benchmark means elsewhere. The
forward test is inert: after the debiasing and whitening of the features every
token has positive cosine to the prototype, so it rejects nothing in
$1{,}000$ of $1{,}000$ episodes and its removal changes no cell. The dilation
is worth $+0.7$ / $+2.1$, but almost exactly because the 2-of-3 per-view
vote is too strict --- relaxing that vote to any-of-3 is worth $+1.2$ /
$+2.0$ on its own --- so the two nearly cancel, and the dilation only
re-inflates what the strict vote eroded. We therefore keep the vote alone,
relaxed, and drop the forward test, the dilation, and its tuned radius. In the
same position the full gate is worse than our vote by $0.5$ / $0.3$~mIoU at
one / five shots over all eight benchmarks (on the four of
\Cref{tab:supp:veto-parts} it is level at five shots), and by $0.3$ and $0.4$ at two and ten
(\Cref{tab:supp:veto-k}); against no veto at all, ours adds $+0.7$, $+1.7$,
$+2.6$ and $+4.7$ at one, two, five and ten shots (\Cref{fig:k-veto}), as the
vote's precision rises from $0.61$ at one shot to $0.77$ at ten ($0.51$ at
two shots, where the majority rule is one of two) while its recall stays at
$0.95$--$0.99$ (medians over episodes).

\begin{table}[h]
  \centering
  \scalebox{0.95}{%
  \begin{tabular}{lcccc}
    \toprule
    Veto & 1-shot & 2-shot & 5-shot & 10-shot \\
    \midrule
    \rowcolor{gray!15}
    none & 58.2 & 60.8 & 62.7 & 65.1 \\
    FROST's gate~\cite{park2026frost} & \underline{58.4} & \underline{62.2} & \underline{65.1} & \underline{69.4} \\
    \rowcolor{red!10}
    backward vote (ours) & \textbf{58.9} & \textbf{62.5} & \textbf{65.4} & \textbf{69.8} \\
    \bottomrule
  \end{tabular}}
  \caption{\textbf{Veto vs.\ shots.} The deployed pipeline with the final mask filtered by nothing, by FROST's full candidate gate, or by our relaxed backward vote (8-benchmark mean).}
  \label{tab:supp:veto-k}
\end{table}

\end{document}